\documentclass[runningheads]{llncs}
\usepackage[T1]{fontenc}
\usepackage{graphicx}
\usepackage{booktabs}
\usepackage[misc]{ifsym}

\def\clr#1#{\@clr{#1}}
\def\@clr#1#2{~{\fboxsep0mm\fbox{\colorbox#1{#2}{\phantom{X}}}}}

\def\cssclr#1{{\fboxsep0mm\fbox{\colorbox[HTML]{#1}{\phantom{X}}}}}

\usepackage{xcolor}
\definecolor{lightblue}{rgb}{0.68, 0.85, 0.9}

\usepackage{hyperref}

\begin{document}
\title{Interactive Counterfactual Generation for Univariate Time Series}
\titlerunning{Interactive Counterfactual Generation}
%
\author{Udo Schlegel\inst{1}\orcidID{0000-0002-1825-0097} \and
Julius Rauscher\inst{1}\orcidID{0000-0001-5955-4487} \and
Daniel A. Keim\inst{1}\orcidID{0000-0001-7966-9740}}
\authorrunning{Schlegel et al.}
%
\institute{University of Konstanz, Konstanz, Germany
\email{u.schlegel@uni-konstanz.de}}
\maketitle              
\begin{abstract}
We propose an interactive methodology for generating counterfactual explanations for univariate time series data in classification tasks by leveraging 2D projections and decision boundary maps to tackle interpretability challenges. 
Our approach aims to enhance the transparency and understanding of deep learning models' decision processes. 
The application simplifies the time series data analysis by enabling users to interactively manipulate projected data points, providing intuitive insights through inverse projection techniques. 
By abstracting user interactions with the projected data points rather than the raw time series data, our method facilitates an intuitive generation of counterfactual explanations.
This approach allows for a more straightforward exploration of univariate time series data, enabling users to manipulate data points to comprehend potential outcomes of hypothetical scenarios. 
We validate this method using the ECG5000 benchmark dataset, demonstrating significant improvements in interpretability and user understanding of time series classification. 
The results indicate a promising direction for enhancing explainable AI, with potential applications in various domains requiring transparent and interpretable deep learning models. 
Future work will explore the scalability of this method to multivariate time series data and its integration with other interpretability techniques.
\end{abstract}
\begin{figure*}[h!t]
    \centering
    \includegraphics[clip, trim=0cm 3.4cm 0.0cm 0.0cm, width=1.00\linewidth]{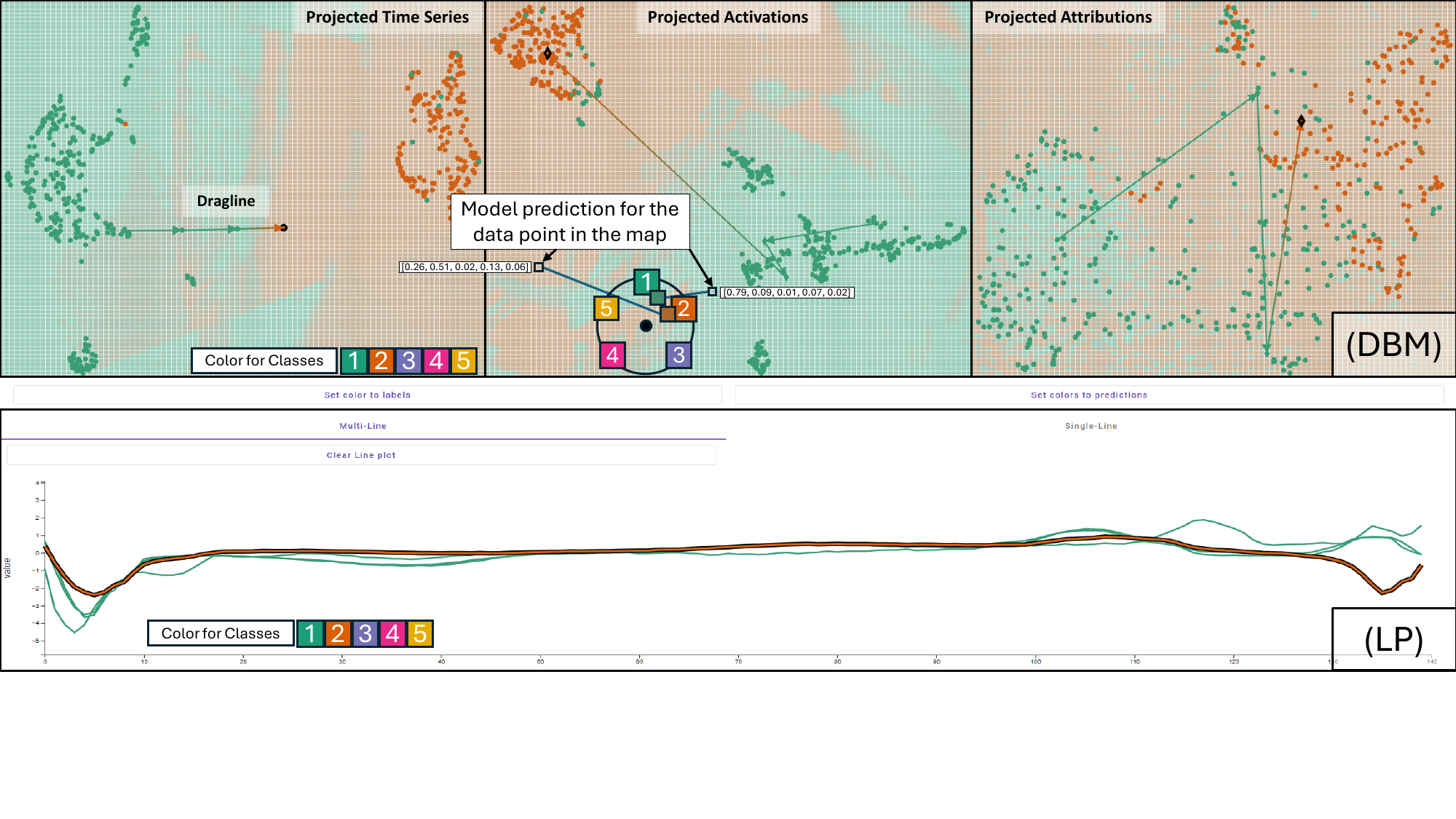}
    \vspace{-2em}
    \caption{Overview of the application: On top, visualizes the decision boundary maps (DBM) of the projections of time series, activations, and attributions of a deep learning time series classifier. The arrows between data points visualize dragged points by a user towards counterfactual explanations. A line plot on the bottom (LP) presents the corresponding time series to the dragged data points in the scatter plot. The highlighted line (upfront with a black stroke) is also highlighted in the scatter plots. The dragline for the points demonstrates interesting patterns in the activations and attributions during a generation of a counterfactual. }
\label{fig:overview}
\end{figure*}

\section{Introduction}



The increasing deployment of complex deep learning models for time series tasks necessitates eXplainable Artificial Intelligence (XAI) methods, which allow humans to understand and interpret the models' decisions. 
This emphasis is particularly crucial in healthcare and predictive maintenance domains, where time series classification plays a pivotal role in predicting outcomes~\cite{mobley_introduction_2002,turbe_interprettime_2022}. 
Analyzing a single-variable (univariate) time series helps us see patterns that change over time, which is crucial for creating understandable and precise models. 
This knowledge is necessary before moving to more complex data, ensuring we fully grasp these time-related patterns as we explore multiple variables (multivariate). 
Thus, the focus on univariate time series facilitates initial simplification and serves as a building block for interpretative modeling in explainable AI applications~\cite{theissler_explainable_2022}.

Deep learning models applied to time series data present a significant challenge in interpretability due to the inherent complexity of the data type.
The sequential nature and the potential for non-linear interdependencies complicate the understanding of how these models make predictions. 
In response to this challenge, XAI endeavors to bridge the gap between complex model behaviors and human interpretability. 
In the context of XAI, activations are the outputs from neurons in a neuronal network model that arise from weighted inputs and bias terms passed through a nonlinear function.
Attributions are a technique that assesses how each input value influences activations, thereby elucidating the model's decision-making process. 
However, visualizing activations or attributions is only part of the solution for users to grasp model decisions completely, as attribution explanations are often hard to interpret~\cite{schlegel_time_2021}.
Another promising avenue within XAI is using counterfactual explanations~\cite{molnar_interpretable_2022}. 
These explanations allow users to explore model decisions by considering alternative scenarios: "What would have happened if X had been different?"~\cite{lipton_contrastive_1990}.
This contrastive approach is particularly valuable as it enables a focus on anomalies or deviations, thereby offering insights into the model's behavior under varied conditions. 
Through such mechanisms, counterfactual explanations demystify the decision-making processes of deep learning time series models and enhance the user's ability to trust and effectively utilize these analytical tools.


In this work, our neural network model-agnostic application provides insights into the model's inner decision-making processes by combining time series data alongside model activations and attributions. 
We propose an interactive application designed to generate counterfactual explanations for univariate deep learning time series models, leveraging projection techniques with inherent inversions, such as Uniform Manifold Approximation and Projection (UMAP)~\cite{mcinnes_umap_2018}, for comprehensive dataset visualization and generation of interactive counterfactuals. 
We further enhance this capability by incorporating optimization techniques that generate time series based on inverse projections in activation and attribution spaces, thus offering an approach to understanding and interpreting model decisions. 
Our application includes interactive line plots, allowing users to dynamically modify time points of time series data, thereby enabling the creation of contrastive explanations in counterfactual scenarios. 
We demonstrate the versatility of our application in the medical domain through its application on the ECG5000 dataset from the UCR benchmark~\cite{dau_ucr_2019}. 
This approach advances the field of explainable AI by making machine learning models more interpretable and serves as a valuable tool for researchers and practitioners looking to gain actionable insights from univariate time series data.


A running demo and the source code is available online at: \\
{\small\href{https://github.com/visual-xai-for-time-series/interactive-counterfactuals-for-time-series}{https://github.com/visual-xai-for-time-series/interactive-counterfactuals-for-time-series}}

\section{Related Work}


Time Series Classification (TSC) is a pivotal task within the broader scope of time series analysis, underpinning numerous applications in fields ranging from finance to healthcare. 
The pursuit of XAI for TSC has unfolded along various dimensions, notably through extracting attributions and counterfactual explanations~\cite{theissler_explainable_2022}.
Attributions highlight the significance of different portions of the time series in the decision-making process of models. 
In contrast, counterfactuals provide insights by presenting hypothetical scenarios with altered model's prediction~\cite{theissler_explainable_2022}.

The generation of counterfactual explanations collected by Guidotti~\cite{guidotti_counterfactual_2022} can be approached through multiple methodologies, each offering unique perspectives, insights, and challenges. 
Despite the advancements in the automatic generation of counterfactuals, concerns have been raised regarding the plausibility of such automatically produced scenarios~\cite{delser_generating_2024}. 
However, in many cases, a general plausibility measure is hard to define as time series data can vary quite task- and domain-dependent, forming different plausible time series for different scenarios.
Del Ser et al.~\cite{delser_generating_2024} highlight the challenges with optimization-based methods such as Wachter et al.~\cite{wachter_counterfactual_2017}, which, while efficient, often yield non-plausible counterfactuals that may not align with real-world possibilities or constraints. 
Delaney et al.~\cite{delaney_instance_2021} propose an approach incorporating a native guide to modify the query time series for a counterfactual, which also can lead to non-plausible time series due to the exchange of segments of the time series and thus cuts at certain points in the time series~\cite{schlegel_visual_2023}. 
These challenges underscore the importance of integrating user interaction with domain knowledge and heuristic approaches in generating counterfactual explanations to generate plausible counterfactual time series.
Our approach incorporated the user in exploring counterfactuals toward already existing time series of a dataset and a model.


Visual analytics has emerged as a crucial component in enhancing the interpretability and accessibility of explainable machine learning, fostering an interactive workflow that bridges the gap between complex models and end users~\cite{sacha_knowledge_2014}. 
As delineated by Spinner et al.~\cite{spinner_explainer_2019}, integrating interactive elements into the analytic process enables users to gain deeper insights into model behavior and decision-making processes. 
Complementing this approach, the What-If Tool proposed by Wexler et al.~\cite{wexler_what_2019} provides a platform for users to engage in perturbation-like sessions, allowing for the exploration of model responses to various hypothetical scenarios.
This hands-on engagement is pivotal in demystifying the often opaque operations of machine learning models.
Our approach includes the user in the generation process of counterfactuals more tightly with the additional help of techniques to enable easier counterfactual generation and thus extends previous methods onto time series.

Schlegel et al.~\cite{schlegel_visual_2023} employ projections to visualize data in intuitive formats and facilitate the generation of counterfactual explanations in line plots. 
Their method emphasizes on attributions and local time series explanations, enabling a focused analysis of individual samples to derive explanations. 
They, however, focus on the interactive change of the time series in line plots, mitigating the possibilities of an inverse projection.
Meanwhile, DECE~\cite{cheng_dece_2020} introduces a visualization system designed explicitly for the exploratory analysis of subgroup counterfactual explanations. 
Although DECE primarily addresses tabular data, its approach exemplifies the potential of visual analytics to make machine learning models more comprehensible and user-friendly. 
Thus, as related work reveals, there is a need for interactive probing of models towards counterfactual generation.
Our approach closes this gap by providing techniques to interactively generate counterfactuals while still enabling users to probe the model concerning, e.g., training data to enable a comparison of generated counterfactuals to already existing samples.

\section{Interactive Generation of Counterfactuals}

Our proposed interactive generation of counterfactual explanations for time series application, as seen in~\autoref{fig:overview}, consists of three visualization components (three decision boundary plots, two line plots) and a preprocessing step to transform the data and model relationship beforehand.
Thus, we introduce the preprocessing first and then describe the interactive visualization components.

\subsection{Preparing the data and the model}

We begin by selecting a time series dataset as the foundation for our preprocessing. 
Next, we train a deep learning model on the selected dataset, utilizing its sophisticated architecture to capture the patterns within the time series. 
Once trained, the model generates predictions for the entire dataset, providing a comprehensive overview of its performance. 
We extract activations from a selected model layer to delve deeper into the model's decision-making process by again predicting the entire dataset and collecting the activations for each sample. 
These activations reveal the internal representations that the model learned for each data sample~\cite{zeiler_visualizing_2014}. 
Additionally, we extract attributions from the model by applying an attribution technique such as DeepLIFT~\cite{shrikumar_learning_2017} similar to Schlegel et al.~\cite{schlegel_towards_2019}, identifying the specific features within the data that impact the model's predictions. 
This dual extraction of activations and attributions offers a nuanced understanding of how the model interprets and reacts to the input data, highlighting the features it deems most important for making predictions.


The next phase of our analysis involves employing Uniform Manifold Approximation and Projection (UMAP)~\cite{mcinnes_umap_2018} to project the time series data, activations, and attributions into a 2D space for visualization purposes through scatter plots. 
This projection facilitates an intuitive understanding of the data's underlying structure as similar samples based on their neighborhood are grouped together in the lower dimensional space. 
We apply UMAP as it provides an inverse projection technique that is inherently due to the map properties it focuses on implementing.
However, other projection techniques with inverse projection capabilities would be possible, e.g., using an autoencoder.
We also generate sampled 2D grid data on the projected space to create decision boundary maps~\cite{espadoto_visual_2019}, offering a unique perspective on data distribution and enhancing the plots. 
The inverse projection of this grid data to time series is approached differently for each component: the time series data is straightforwardly inverse projected, while activations and attributions require other techniques. 
An activation maximization approach is employed to transform the activations back into time series data by not maximally activating the selected neuron but towards the wanted activation. 
We adapt the general activation maximization algorithm to first compare the activations of a randomly generated input time series to our grid-generated activations.
We calculate the gradients of the input based on the resulting loss of the comparison and use these gradients for a gradient ascent to change the input time series.
Through a few steps, we approximate the corresponding time series for the selected activation.
A depiction of the process can be seen in~\autoref{fig:time-series-from-activation} initialized with a click on the scatter plot.
The strategy is also applied to attributions, with the attribution technique generating attributions for each intermediate step of the time series. 
The generation process for the attributions can be time-consuming but can be parallelized to expedite the process. 
However, for small granular grids or large value ranges, this can result in a high computational load for both activations and attributions.

\subsection{Interactive Visualizations Workspace}
Following the principles of visual analytics, we rely on multiple interactive linked views that provide different perspectives to support the user in generating and investigating counterfactual explanations.

By employing projection decision boundary maps (DBM), we can make time series data more accessible, allowing users to grasp temporal patterns at a glance. 
The incorporation of decision density maps within the projection space offers a tangible representation of how decisions are distributed across different ranges.
Allowing the user to drag individual projected time series instances further enhances the projection space and empowers the user to inspect how alterations in data impact model outcomes. 
These principles are employed in separate projections for the time series (left), the activations (middle), and the attributions (right in the DBM of~\autoref{fig:overview}), allowing the user to explore the model on different semantic levels.

The application also integrates line plots as a more conventional visualization technique for time series analysis.
This includes multi-line plots for comparative analysis and single-line plots with interactive capabilities, facilitating the generation of counterfactual explanations by permitting users to experiment with modifications and immediately observe the effects. 


\subsubsection{Projection Decision Boundary Maps}



Each time series, activation, or attribution is transformed into a 2D sample point, utilizing UMAP to project data from n dimensions into a 2D space, with its color denoting either the ground truth label or a predicted label derived from the model. 
This method simplifies the complex dimensionality of time series data and facilitates an intuitive understanding of the model's accuracy and biases. 
The backdrop of this visualization features a decision map, employing a dense pixel approach as described by Rodrigues et al.~\cite{rodrigues_constructing_2019}.
This technique ensures that each pixel or region within the map reflects the model's decision for the data projected onto that specific area. 
Since deep learning model outputs are typically probabilistic if a softmax is applied, the color within this density map is adjusted to represent class color weightings. 
For instance, in a scenario involving five classes, each class is assigned a color of a color map. 
The model's predictions, such as an output of $(0.5, 0.2, 0.1, 0.1, 0.1)$ and a color scale of \cssclr{1b9e77}\cssclr{d95f02}\cssclr{7570b3}\cssclr{e7298a}\cssclr{e6ab02} (Dark2~\cite{harrower_colorbrewer_2003}), lead to visually blending colors to class probabilities, such as~\cssclr{66825f}, offering a visually intuitive representation of prediction confidence and class distribution.

\begin{figure}[h!t]
    \centering
    \includegraphics[clip, trim=0.0cm 15.3cm 22.6cm 0.0cm, width=0.99\linewidth]{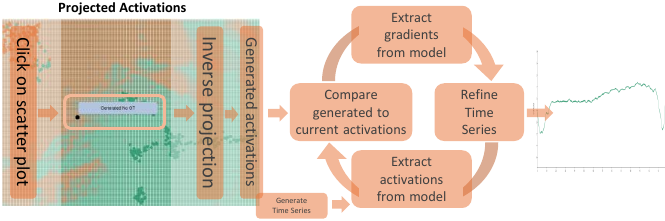}
    \vspace{-1em}
    \caption{Based on a click on the projected activations, the inverse projections generate activations. These activations are used in an optimization loop that takes a generated time series and refines it based on the difference between the current time series activations of the model and the inverse-generated activation.}
    \label{fig:time-series-from-activation}
    \vspace{-1em}
\end{figure}

Hovering over data points highlights those points within the scatter plots and emphasizes corresponding data in line plots, fostering a cohesive exploration across visualizations. 
Also, hovering shows a tooltip with the ground truth, the prediction, and the prediction probabilities.
Clicking on a data point integrates it into line plots for more detailed comparison or analysis. 
Dragging a point offers updates in the scatter and line plots, including the model's re-prediction for the adjusted data point with an innovative feature that allows for the inverse projection of a dragged point into a region without other data points such as depicted in~\autoref{fig:time-series-from-activation}.
The original point and the new dragged point get connected by a line with a color gradient, demonstrating the change in prediction if it happened.
Furthermore, the same feature allows the inverse projection of a clicked area in the scatter plot's free space, automatically generating a hypothetical time series as seen in~\autoref{fig:time-series-from-activation}. 
This generated series is then depicted in the line and scatter plots, bridging the gap between high-dimensional data representation and tangible, actionable insights as seen in~\autoref{fig:overview}. 
Such interactions empower users to dynamically manipulate and interrogate the data and facilitate a deeper understanding of the model's behavior and underlying data.

\subsubsection{Comparison Multi Line Plot}


The multi-line plot feature offers users a traditional visualization for analyzing time series data, seamlessly integrating with interactive scatter plots to enhance data exploration. 
Users can add lines to the multi-line plot directly from scatter plots, with each line representing either existing data points or newly generated ones. 
This integration facilitates an intuitive comparison and analysis of time series data. 
The color coding of each line corresponds with the model's predictions, providing immediate visual feedback on the data's classification output. 
Additionally, an interactive element is introduced whereby hovering over a line in the multi-line plot automatically highlights the corresponding data point in the scatter plots, linking different visualizations and aiding in identifying patterns or anomalies. 
A clear button is included to provide users with the convenience of resetting the line plot, removing all lines, and allowing for a fresh start in the data exploration.
This feature is indispensable for users who wish to reset their analysis and start anew, ensuring that the visualization space remains manageable and conducive to insightful exploration without overplotting.

\subsubsection{Counterfactual Generation Line Plot}


\begin{figure*}[h!t]
    \centering
    \includegraphics[clip, trim=0.0cm 8.4cm 3.9cm 0.0cm, width=1.0\textwidth]{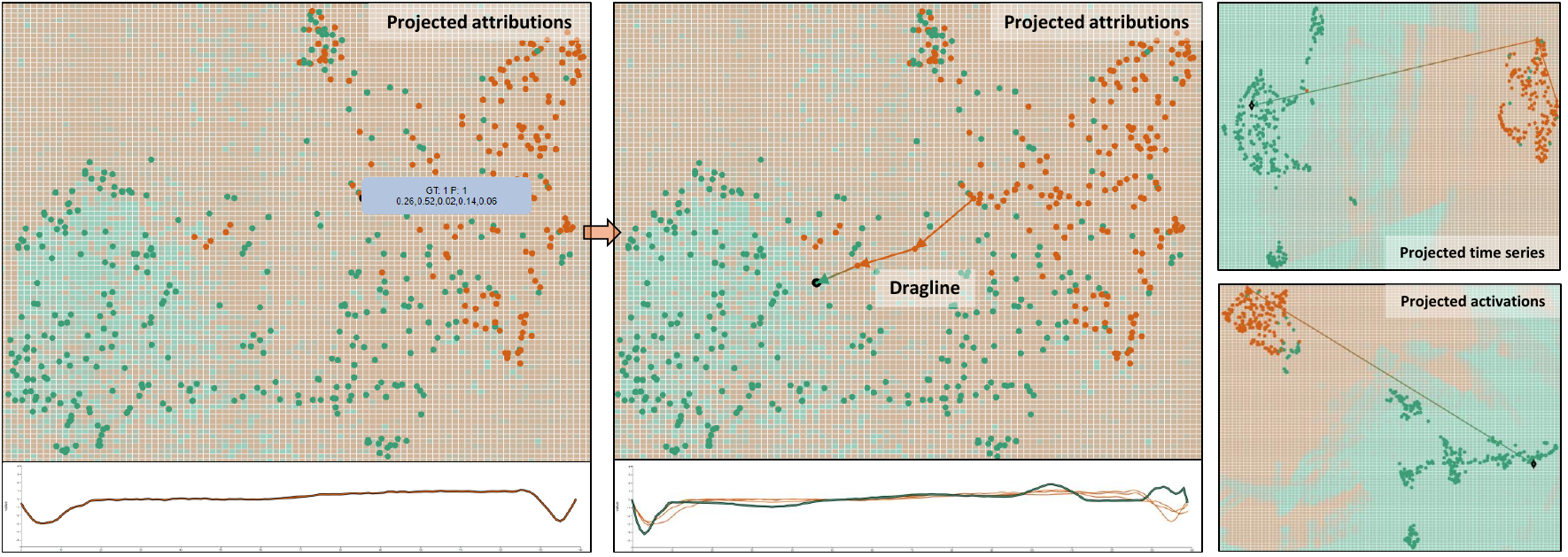}
    \vspace{-2em}
    \caption{Generating counterfactual explanations based on the projected attributions by slowly dragging a data point to a region with another class prediction on the dense decision map on the projected attributions. Generated time series seem plausible in the projected time series, and projected activations scatter plots never come close to borders and stay in regions with other data points.}
    \label{fig:counterfactual-attributions}
    \vspace{-1em}
\end{figure*}


In complement to the multi-line visualization, another line plot feature is designed for more granular interaction, focusing on modifying a single line at a time. 
This plot restricts the display to one line only, necessitating the clearing of the existing line before a new one can be introduced. 
Mirroring the multi-line approach, the color of the singular line reflects the model's prediction, providing an immediate visual cue to the user about the classified outcome or regression value. 
Uniquely, this plot enhances user interaction by plotting each time point in the time series as a draggable point along the line, as seen in~\autoref{fig:counterfactual-attributions}. 
This allows users to adjust the time series data in the y-direction, facilitating hands-on exploration of data manipulation effects similar to Schlegel et al.~\cite{schlegel_visual_2023}.

A significant addition to this feature is plotting the original time series line in the background as a reference point for any modifications made. 
Once applied, such changes to the time series line can be projected back into the scatter plots. 
This projection lets users visually assess how their modifications impact the time series distribution and the underlying model's decisions, activations, and attributions. 
This interactive process not only aids in understanding the direct implications of data adjustments on model predictions but also provides insights into the complex dynamics of the model's internal decision-making mechanisms, offering a deeper exploration into the interpretability of models through visual analytics.

\section{Model and Data Exploration on Use Cases}


This section delves into the practical applications of our chosen benchmark model and dataset to demonstrate specific use cases. These examples illustrate the capabilities of the established models and showcase the dataset's versatility in real-world scenarios. 
Furthermore, an online demo is available for those interested in exploring a wider array of models and datasets, offering a hands-on experience with additional resources.


\textbf{Users --}
According to Spinner et al. ~\cite{spinner_explainer_2019}, there are three user groups in XAI: model novices, model users, and model developers with different knowledge of deep learning models.
Our primary audience includes model developers and model users, each with distinct needs and objectives. 
These individuals seek not only to understand the underlying mechanics of models but also to engage in debugging activities and to test these models against unseen data and unknown distribution shifts. 
We employ a contrastive explanation process that utilizes counterfactuals to address these requirements effectively. 
This method explores how different inputs can lead to varying outcomes, thus providing deeper insights into model behavior and enhancing a robustness analysis under varied conditions.



\textbf{Model and datset --}
The ECG5000 dataset~\cite{dau_ucr_2019}, a crucial resource for time series analysis in the medical domain, comprises 5,000 segments of electrocardiogram (ECG) data. 
This dataset is split into 500 segments for training and 4,500 segments for testing purposes, with each segment encapsulating 140 time points of ECG readings. 
This dataset is categorized into five classes, with a significant imbalance among them, as evidenced by the distribution in the training set: 292 instances in class 1, followed by 177 in class 2, 19 in class 3, 10 in class 4, and a mere 2 in class 5. 
A Conv1D ResNet architecture comprising three ResNet Blocks and a concluding linear layer for classification was employed to tackle the challenges of such an imbalanced dataset. 
This choice is motivated by ResNet's proven efficacy in time series data, thanks to its ability to capture deep temporal dependencies~\cite{ismailfawaz_deep_2019}. 
The model's performance on the ECG5000 dataset attests to the effectiveness of ResNet in this context, achieving an accuracy of 90.6\% (training) and 89.33\% (test). 




\textbf{Task 1: Exploration of data and model decisions relationship --} 
Our application provides a robust way for dissecting the intrinsic dynamics between datasets and model decisions, showcased through decision boundary maps using UMAP projections and dragging interactions. 
Users can hover over sample points in the maps to delve into data points and their impact on model outcomes, observing tooltips and the underlying decision maps while also seeing other highlighted points across various plots. 
These maps clearly define the model’s decision boundaries at multiple levels - data, activations, and attributions - offering a comprehensive insight into internal processes. 
This feature, coupled with the inverse projection that populates unseen data spaces, provides deeper insights into the model's behavior towards novel or hypothetical time series data. 
Further scrutiny of these plots, as detailed in~\autoref{fig:overview}, exposes a tendency of the model to predict major classes accurately, revealed through the dominance of two colors in the attribution maps that indicate a biased distribution. 
The use of 2D UMAP projection offers an essential overview of model predictions but shows limited impressiveness when actual labels are applied, reflecting similar patterns to those observed in the ResNet model for majority classes. 
This analysis is vital for understanding class imbalance issues and determining that the model predominantly underfits minority classes, stressing the necessity to refine model training to enhance predictive accuracy across all classes.




\textbf{Task 2: Interactive generation of counterfactual explanations using inverse projections --}
Our application enhances the generation of counterfactual explanations through inverse projections, allowing interactive manipulation of data points within scatter plots to create new time series data either by clicking in empty spaces or dragging data points across different class backgrounds on the decision map. 
These modifications are immediately visible in line plots, providing an intuitive means for exploring changes. 
Users can further refine adjustments using the single-line plot feature to modify time points within the time series and directly observe effects on model predictions, offering a streamlined toolkit for constructing and comprehending counterfactual scenarios. 
In practice, selecting and gradually dragging a point within a cluster or from an arbitrary location towards the decision boundaries, as depicted in~\autoref{fig:overview} and~\autoref{fig:counterfactual-attributions}, provides significant insights. When directed towards another cluster, this technique leverages inverse projection to generate plausible new time series counterfactual explanations. 
Although this method maintains high plausibility within the time series projection domain, it proves less effective when applied to projected activations, often resulting in the generation of time series for minority classes with incorrect predictions and pinpointing areas for model enhancement. 
In the context of projected attributions, while the dragging technique effectively generates new data points, it is challenging due to the narrow region delineated for one of the majority classes, indicating a complex landscape of model decisions where the utility of dragging for counterfactual generation varies across different projection types, each providing unique insights into the model's decision-making processes.

\begin{figure*}[h!tb]
    \centering
    \includegraphics[clip, trim=0cm 3.4cm 0.0cm 0.0cm, width=1.00\linewidth]{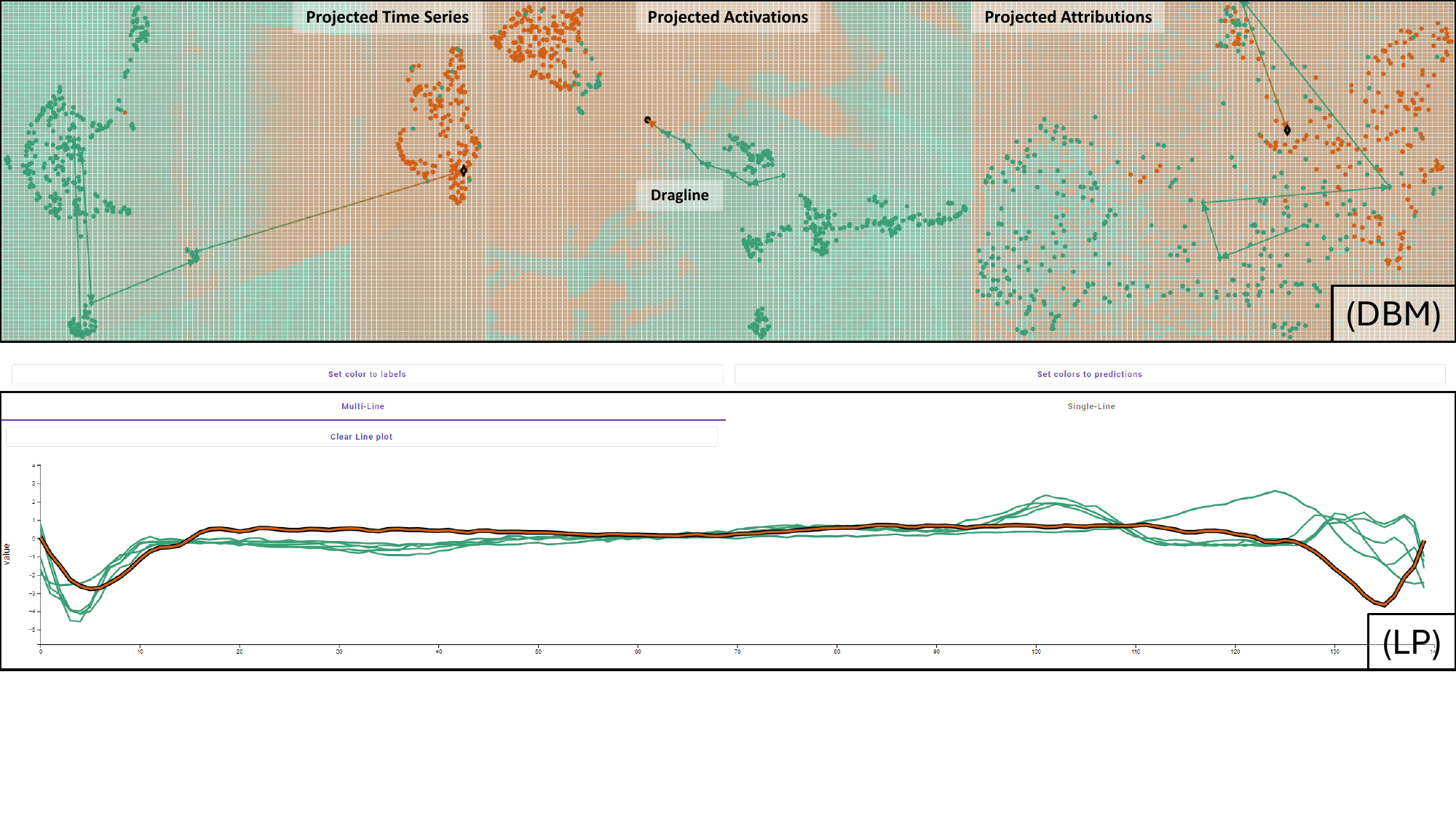}
    \vspace{-2em}
    \caption{On top, visualizes the decision boundary maps (DBM) of the projections of time series, activations, and attributions of a deep learning time series classifier. The arrows between data points visualize dragged points by a user towards counterfactual explanations in the activations. A line plot on the bottom (LP) presents the corresponding time series to the dragged data points in the scatter plot. The highlighted line (upfront with a black stroke) is also highlighted in the scatter plots. The dragline for the points demonstrates interesting patterns in the original time series and attributions during a generation of a counterfactual in the activations. Especially interesting is the attribution DBM, as the generated time series jumps around quite heavily.}
\label{fig:dragline-activations}
\end{figure*}

~\autoref{fig:dragline-activations} visualizes the process of generating counterfactuals using activations, thereby facilitating easier access to decision boundaries. 
However, dragging a selected time series sample to a decision boundary may sometimes require a step beyond the border. 
This necessity might arise from our methodology for generating the time series, as the gradient ascent employs a nearest neighbor approach and could potentially reach a local minimum, necessitating a further leap over the decision boundary to generate a counterfactual.
Nevertheless, the resulting time series appears promising. 
More intriguing are the projected time series and projected attributions plots, which reveal interesting jumps to different locations, with attributions appearing in regions where few other points exist. 
Such behavior could occur if the attributions become implausible or unseen by the projection technique. 
Investigating these scenarios can provide further insights into the attribution and projection techniques. 
In our case, the generated time series are too close to the time series attributions at the border regions of the plot.
The projected time series plot also reveals patterns in our time series generation process, wherein the nearest neighbor approach identifies time series similar to the newly generated ones. 
By introducing some randomness to incorporate artificial noise, the generation process can potentially be enhanced to produce more novel time series. 
Nevertheless, for our exploration of the model's decisions, the current method proves to be quite effective.

\section{Limitations \& Conclusions \& Future Works}



\textbf{Limitations --}
Our application encounters a notable limitation \textit{concerning the plausibility} of generated counterfactual time series, emphasized by Del Ser et al.~\cite{delser_generating_2024}. 
Specifically, the reliance on UMAP for the projection of time series data inherently challenges the plausibility of counterfactuals when such data points are located on the projections' edges, signaling a likelihood of implausibility. 
Furthermore, the stochastic nature of UMAP underscores the difficulty in achieving deterministic projections and also inverse projections. 
To mitigate this, we suggest the incorporation of more deterministic methods, such as inverse Neural Network projections (iNN) as proposed by Espadoto et al.~\cite{espadoto_visual_2019}, and integrating interactive modifications in line plots as recommended by Schlegel et al.~\cite{schlegel_visual_2023}. 
These enhancements aim to refine the generation process, improving the plausibility and relevance of counterfactuals within the confines of our interactive application and the user experience.
Another factor is the \textit{color space} used to generate the background for the DBM.
Due to many similar predictions toward the majority class and the overall color space generation, a clear border between the different classes and even between the different probabilities is hard to observe.
Changing the color space in a more focused way to something with a larger variety of colors and a clever combination can facilitate the discovery of patterns more easily, similar to El-Assady et al.~\cite{elassady_semantic_2022}.
Also, our current approach \textit{only supports univariate time series} due to, e.g., using single line plots to visualize the generated counterfactuals and original data.
Multivariate time series would need a more sophisticated visualization to enable users to work with the data.

\textbf{Conclusion --}
In conclusion, our interactive application designed to generate counterfactual explanations for time series classification models represents a step forward in employing advanced visualization techniques, such as inverse projections via UMAP and optimization steps derived from activations or attributions. 
As demonstrated by our use cases, these techniques support the generation process of counterfactuals and potentially lead to more plausible and insightful outcomes when coupled with user interactions. 
Furthermore, the application of projections serves as a valuable tool for evaluating the plausibility of counterfactuals not only in the original time series but also in the associated activations and attributions. 
By offering a visual means to identify when a counterfactual might fall outside the expected distribution, our approach provides a mechanism for discerning the feasibility and relevance of these hypothetical scenarios, enhancing the interpretability and applicability of time series classification models in various domains.

\textbf{Future Work --}
Future work could extend our current univariate time series approach to accommodate multivariate data. 
The inverse projection methods, such as UMAP~\cite{mcinnes_umap_2018}, are capable of projecting higher-dimensional data into 2D spaces, indicating the potential for such adaptation. 
However, this process typically results in information loss, potentially degrading reconstruction accuracy. 
Addressing this challenge could involve \textit{employing varied projections across different dimensionalities} and interlinking these latent spaces to enable upscaling from a 2D to a higher-dimensional latent space, such as 100 dimensions, and then reconstructing the time series.
Additionally, exploring \textit{alternative inverse projection techniques} like iNN~\cite{espadoto_visual_2019} may help mitigate these issues, as these methods are not confined to specific manifolds, unlike UMAP~\cite{mcinnes_umap_2018}. 
The core optimization strategy for deriving time series based on activations or attributions remains applicable, facilitating the generation of counterfactuals for these measures.
The transition from univariate to multivariate time series is primarily hindered by the challenges associated with inverse projection and the consequent increase in information and dimensionality.
Further research could also explore \textit{other visualization techniques to enhance user comprehension}, particularly for those with limited domain knowledge or AI expertise. 
For instance, incorporating the exploration approach by Schlegel et al.~\cite{schlegel_visual_2023} could break down the data and model information even further, making it more accessible for non-experts.
Additionally, incorporating further visualizations can be enhanced by \textit{including verbal explanations} to provide a more comprehensive understanding of the model's decisions, thereby supporting a wider user base~\cite{sevastjanova_going_2018}.
This verbalization may also enhance explanations in cases of heavily overplotted visualizations where the UMAP 2D projection does not perform as intended.

\begin{credits}
\subsubsection{\ackname} 
This work has been partially supported by the Federal Ministry of Education and Research (BMBF) in VIKING (13N16242).
\end{credits}
%
%
%
\bibliographystyle{splncs04}
\bibliography{main}

\end{document}